\documentclass[runningheads]{llncs}
\usepackage[T1]{fontenc}
\usepackage{graphicx}
\usepackage[misc,geometry]{ifsym}
\usepackage{amsmath}
\usepackage{color}
\begin{document}
\title{Unsupervised Cross-Modality Domain Adaptation for Vestibular Schwannoma Segmentation and Koos Grade Prediction based on Semi-Supervised Contrastive Learning}

\titlerunning{Unsupervised Cross-Modality Domain Adaptation}
%
\author{Luyi Han\inst{1,2} \and
Yunzhi Huang\inst{3}\thanks{Luyi Han and Yunzhi Huang contributed equally to this work.} \and
Tao Tan\inst{2}\textsuperscript{(\Letter)} \and
Ritse Mann\inst{1,2} 
}
\authorrunning{L. Han et al.}
%
\institute{Department of Radiology and Nuclear Medicine, Radboud University Medical Center, Geert Grooteplein 10, 6525 GA, Nijmegen, The Netherlands.\\
\and
Department of Radiology, The Netherlands Cancer Institute, Plesmanlaan 121, 1066 CX, Amsterdam, The Netherlands.\\
\and
School of Automation, Nanjing University of Information Science and Technology, Nanjing 210044, China.\\
\email{\{taotanjs\}@gmail.com}
}
\maketitle              
\begin{abstract}
Domain adaptation has been widely adopted to transfer styles across multi-vendors and multi-centers, as well as to complement the missing modalities. In this challenge, we proposed an unsupervised domain adaptation framework for cross-modality vestibular schwannoma (VS) and cochlea segmentation and Koos grade prediction.
We learn the shared representation from both ceT1 and hrT2 images and recover another modality from the latent representation, and we also utilize proxy tasks of VS segmentation and brain parcellation to restrict the consistency of image structures in domain adaptation. After generating missing modalities, the nnU-Net model is utilized for VS and cochlea segmentation, while a semi-supervised contrastive learning pre-train approach is employed to improve the model performance for Koos grade prediction.
On CrossMoDA validation phase Leaderboard, our method received rank 4 in task1 with a mean Dice score of 0.8394 and rank 2 in task2 with Macro-Average Mean Square Error of 0.3941. Our code is available at \url{https://github.com/fiy2W/cmda2022.superpolymerization}.

\end{abstract}
\section{Introduction}
Domain adaptation has recently been employed in various clinical settings to improve the applicability of deep learning approaches.
The goal of Cross-Modality Domain Adaptation (CrossMoDA) challenge~\footnote{\url{https://crossmoda2022.grand-challenge.org/}} is to segment two key brain structures, namely vestibular schwannoma (VS) and cochlea. It also requires predicting the Koos grading scale for VS.
The two tasks are required for the measurement of VS growth and evaluation of the treatment plan (surveillance, radiosurgery, open surgery).
Although contrast-enhanced T1 (ceT1) MR imaging is commonly used for patients with VS in diagnosis and surveillance, the research on non-contrast imaging, such as high-resolution T2 (hrT2), is growing due to lower risk and more efficient cost.
Therefore, CrossMoDA aims to transfer the model learned from annotated ceT1 images to unpaired and unlabeled hrT2 images based on domain adaptation.

\section{Related Work}
Unsupervised domain adaptation for VS and cochlea segmentation has been extensively validated in previous research~\cite{dorent2022crossmoda}. Most of them employ an image-to-image translation method, \emph{e.g.} CycleGAN~\cite{zhu2017unpaired}, to generate pseudo-target domain images from source domain images. And then generated images and the corresponding manual annotations are used to train the segmentation models.
Dong \textit{et al.}~\cite{dong2021unsupervised} utilize NiceGAN~\cite{chen2020reusing}, which is trained by reusing discriminators for encoding, to improve the performance of domain adaptation and further segmentation.
Choi~\cite{choi2022using} proposes a data augmentation method by halving the intensity in the tumor area for generated hrT2.
Shin \textit{et al.}~\cite{shin2022cosmos} employ an iterable self-training strategy in their method: (1) train the student model with annotated generated hrT2 and pseudo-labeled real hrT2; (2) make the student a new teacher and update the pseudo label for real htT2.
Following these works, our proposed method focuses more on extracting joint representations from multi-modality MRIs, which can reduce the distance between different modalities in the latent space.

Classification task for medical images is always more difficult than segmentation due to fewer annotations. In recent years, contrastive learning has led to state-of-the-art performance in self-supervised representation learning~\cite{wu2018unsupervised,bachman2019learning,he2020momentum,khosla2020supervised}. The key idea is to reduce the distance between an anchor and a “positive” sample in latent space, and distinguish the anchor from other “negative” samples. Based on this, contrastive learning can be applied to multi-modality pretraining. In order to improve the sensor setup flexibility of the robot, Meyer \textit{et al.}~\cite{meyer2020improving} propose a multimodal approach based on contrastive learning to learn from RGB-Depth images. Yuan \textit{et al.}~\cite{yuan2021multimodal} develop an approach for joint visual-textural pretraining that focuses on both intra-modality and inter-modality learning. For medical image analysis, Huang \textit{et al.}~\cite{huang2021gloria} develop an attentional contrastive learning framework for global and local representation learning between images and radiology reports. Inspired by these works, contrastive learning is employed at the pretraining phase to mine multi-modality representation strategically for different types of samples.

\begin{figure}
    \centering
    \includegraphics[width=1\linewidth]{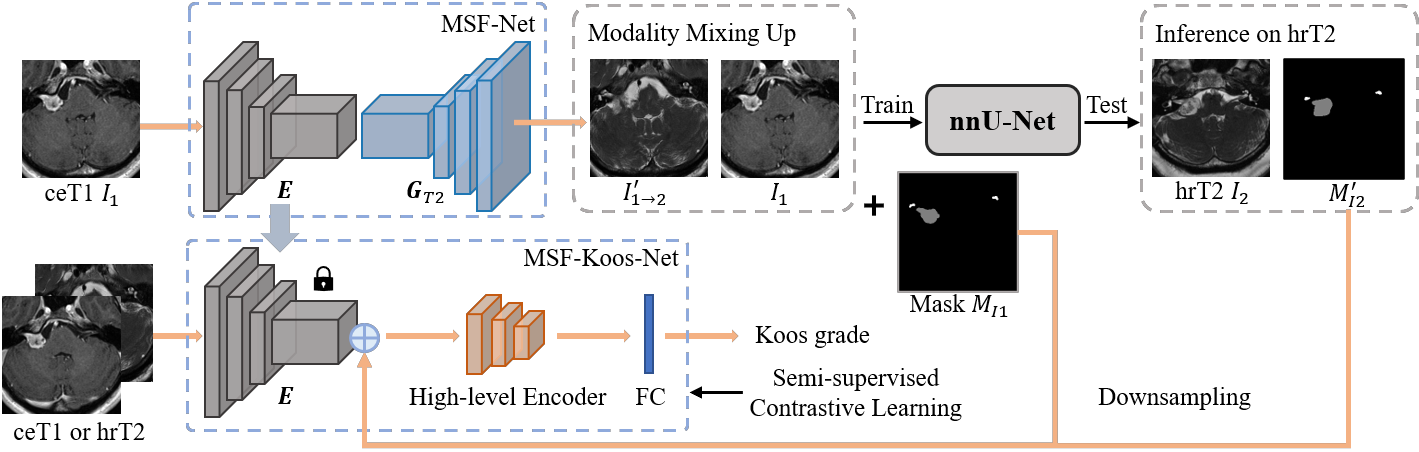}
    \caption{Overview of the proposed unsupervised domain adaptation segmentation and classification framework.}
    \label{fig:framework}
\end{figure}

\section{Method}
\subsection{Framework Overview}
Figure~\ref{fig:framework} illustrates the proposed unsupervised domain adaptation segmentation and
classification framework. We first employ a multi-sequence fusion network (MSF-Net) to generate the corresponding hrT2 image from a given ceT1 image.
To train a robust segmentation network, we pool real ceT1 images and generated hrT2 images together, instead of pairing them, to ensure the nnU-Net~\cite{isensee2021nnu} model is able to predict ceT1 and hrT2 images blindly.
By leveraging the predicted segmentation mask and pre-trained MSF-Net, we propose MSF-Koos-Net based on semi-supervised contrastive learning to predict Koos grade.

\begin{figure}
    \centering
    \includegraphics[width=1\linewidth]{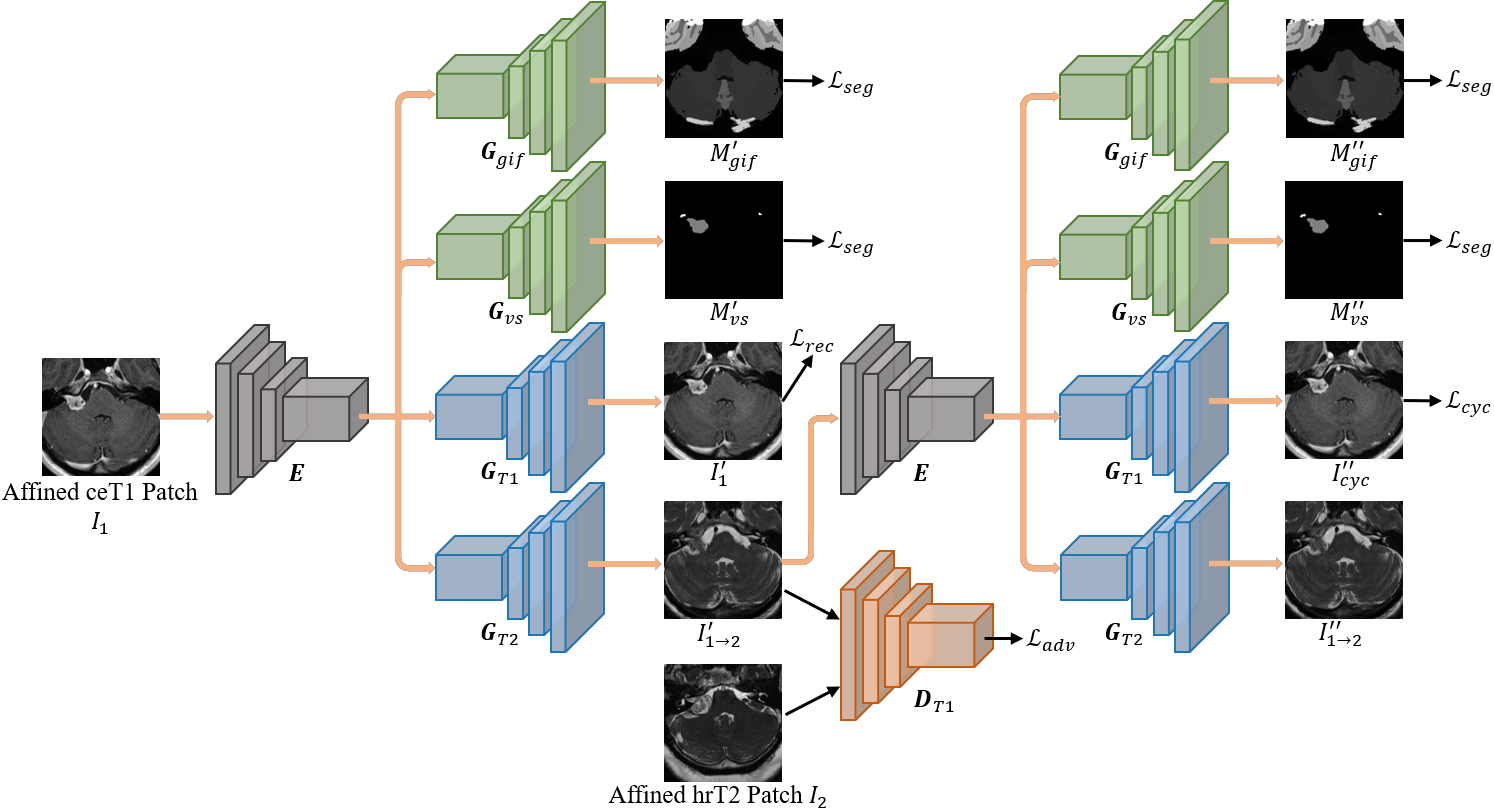}
    \caption{The architecture of MSF-Net. The reverse transform direction (from real hrT2 to fake ceT1) is omitted for ease of illustration. Not that, both directions share weights for the model, and no proxy paths ($\mathbf{G}_{vs}$ and $\mathbf{G}_{gif}$) are involved in the reverse direction due to lack of annotations.}
    \label{fig:architecture}
\end{figure}

\subsection{Cochlea and VS segmentation based on unsupervised domain adaptation}
Figure~\ref{fig:architecture} illustrated the architecture of the proposed MSF-Net. Although ceT1 and hrT2 MRI images differ in image resolution and appearance, organs' representations from an identical subject are commonly embedded in the latent space. Based on this, we employed a share-weighted encoder $\mathbf{E}$ to extract the domain-free representations from both ceT1 and hrT2 images. Then, two decoders ($\mathbf{G}_{T1}$ and $\mathbf{G}_{T2}$) was constructed to recover the different parametric MRI sequences from the latent representations. The reconstruction loss is as follows,
\begin{equation}
    \label{eq:reconstruction}
    \begin{aligned}
        \mathcal{L}_{rec}=\lambda_{r}\cdot(\|I'_1-I_1\|_1 + \|I'_2-I_2\|_1) + \lambda_{p}\cdot(\mathcal{L}_{p}(I'_1, I_1) + \mathcal{L}_{p}(I'_2, I_2))
    \end{aligned}
\end{equation}
where $I'_1=\mathbf{G}_{T1}(\mathbf{E}(I_1))$, $I'_2=\mathbf{G}_{T2}(\mathbf{E}(I_2))$, $\| \cdot \|_1$ is a $L_1$ loss, and $\mathcal{L}_{p}$ refers to the perceptual loss based on pre-trained VGG19. $\lambda_{r}$ and $\lambda_{p}$ are weight terms and are set to be $10$ and $0.01$.

Inspired by Cycle-GAN~\cite{zhu2017unpaired}, we utilize the adversarial loss to achieve the domain adaptation and employ cycle consistency loss to force the consistency of the anatomical structures.
\begin{equation}
    \label{eq:adversarial}
    \begin{aligned}
        \min_{\mathbf{D}_{T1},\mathbf{D}_{T2}}\max_{\mathbf{G}}\mathcal{L}_{adv}&= \|\mathbf{D}_{T1}(I_1)-1\|_2+\|\mathbf{D}_{T1}(I'_{2\rightarrow 1})\|_2 \\ &+ \|\mathbf{D}_{T2}(I_2)-1\|_2+\|\mathbf{D}_{T2}(I'_{1\rightarrow 2})\|_2
    \end{aligned}
\end{equation}
\begin{equation}
    \label{eq:cycle}
    \begin{aligned}
        \mathcal{L}_{cyc}=\|I''_{1\rightarrow 2\rightarrow 1}-I_1\|_1+\|I''_{2\rightarrow 1\rightarrow 2}-I_2\|_1
    \end{aligned}
\end{equation}
where $I'_{1\rightarrow 2}=\mathbf{G}_{T2}(\mathbf{E}(I_1))$, $I''_{1\rightarrow 2\rightarrow 1}=\mathbf{G}_{T1}(\mathbf{E}(I'_{1\rightarrow 2}))$, $I'_{2\rightarrow 1}$ and $I'_{2\rightarrow 1\rightarrow 2}$ are formulated similarly, $\| \cdot \|_2$ is a $L_2$ loss.

To further restrict the image structure during domain adaptation, especially for tumors, we employ two proxy tasks for real ceT1 images, including VS segmentation ($\mathbf{G}_{vs}$) and brain parcellation ($\mathbf{G}_{gif}$) whose labels are obtained with the Geodesic Information Flows (GIF) algorithm.
\begin{equation}
    \label{eq:segmentation}
    \begin{aligned}
        \mathcal{L}_{seg}&=\mathcal{L}_{ce}(M'_{vs}, M_{vs}) + \mathcal{L}_{dsc}(M'_{vs}, M_{vs}) \\
        &+ \mathcal{L}_{ce}(M'_{gif}, M_{gif}) + \mathcal{L}_{dsc}(M'_{gif}, M_{gif})
    \end{aligned}
\end{equation}
where $\mathcal{L}_{ce}$ refers to the cross entropy loss and $\mathcal{L}_{dsc}$ indicates the dice similarity coefficient loss.

\subsection{Koos grade prediction based on semi-supervised contrastive learning}
\begin{figure}
    \centering
    \includegraphics[width=1\linewidth]{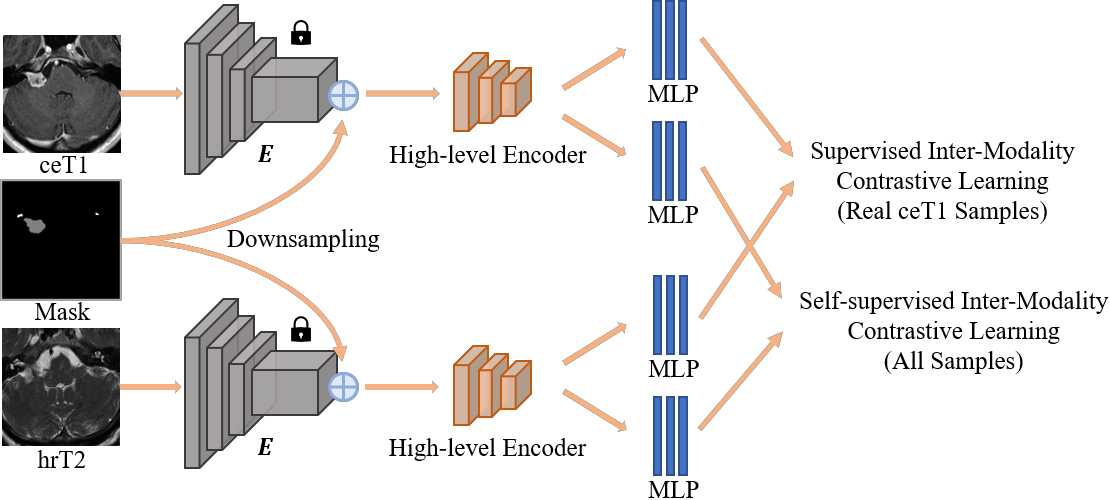}
    \caption{The architecture of MSF-Koos-Net.}
    \label{fig:classification}
\end{figure}

Figure~\ref{fig:classification} illustrates the architecture of MSF-Koos-Net. The frozen pre-trained encoder $\mathbf{E}$ from MSF-Net is employed to extract low-level features from both ceT1 and hrT2 images. To pay more attention to the tumor region, we concatenate the low-level image features with the predicted segmentation mask of the tumor. Then followed with a high-level encoder $\mathbf{E}_\mathcal{H}$ to extract high dimension features and a full connection layer to output the predicted Koos grade.
To achieve better performance of Koos grade prediction with limited data, both supervised and self-supervised contrastive learning~\cite{khosla2020supervised} are utilized to pre-train the MSF-Koos-Net.

\subsubsection{Self-supervised contrastive learning.}
After generating the missing modality, a dataset that includes paired ceT1 and hrT2 is composed. Within a multi-modality batch, let $\mathcal{D}$ be the group of indexes for these samples. For self-supervised contrastive learning, only cross-modality samples with the same indexes as the source sample are positive. The loss function is defined as follows,
\begin{equation}
    \label{eq:self-contrast}
    \begin{aligned}
        \mathcal{L}_{self} = -\sum_{i\in\mathcal{D}}\log\frac{\exp{(z^{(i)}_1\cdot z^{(i)}_2/\tau)}}{\sum_{j\in\mathcal{D}}\exp{(z^{(i)}_1\cdot z^{(j)}_2/\tau)}} \cdot \frac{\exp{(z^{(i)}_1\cdot z^{(i)}_2/\tau)}}{\sum_{j\in\mathcal{D}}\exp{(z^{(j)}_1\cdot z^{(i)}_2/\tau)}}
    \end{aligned}
\end{equation}
where $z_{1,2}=\mathcal{F}_{self}(\mathbf{E}_\mathcal{H}(\mathbf{E}(I_{1,2}))$ refers to features extracted from ceT1 and hrT2 images, $\mathcal{F}_{self}$ indicates the projection network for self-supervised contrastive learning, the $\cdot$ symbol denotes Scalar Product, $\tau$ refers to the scalar temperature parameter.

\subsubsection{Supervised contrastive learning.}
To leverage Koos grade for pretraining, supervised contrastive learning is employed to enlarge the inter-grade difference and intra-grade similarity. We use samples from real ceT1 and the corresponding fake hrT2 pairs, and let $\mathcal{A}$ be the index group for the annotated samples in a multi-modality batch. The loss takes the following form,
\begin{equation}
    \label{eq:sup-contrast}
    \begin{aligned}
        \mathcal{L}_{sup} = -\sum_{i\in\mathcal{A}}&\frac{1}{|\mathcal{P}(i)|}\sum_{p\in\mathcal{P}(i)} \\
        &\log\frac{\exp{(q^{(i)}_1\cdot q^{(p)}_2/\tau)}}{\sum_{j\in\mathcal{A}}\exp{(q^{(i)}_1\cdot q^{(j)}_2/\tau)}} \cdot \frac{\exp{(q^{(p)}_1\cdot q^{(i)}_2/\tau)}}{\sum_{j\in\mathcal{A}}\exp{(q^{(j)}_1\cdot q^{(i)}_2/\tau)}}
    \end{aligned}
\end{equation}
where $q_{1,2}=\mathcal{F}_{sup}(\mathbf{E}_\mathcal{H}(\mathbf{E}(I_{1,2}))$ refers to features extracted from ceT1 and hrT2 images, $\mathcal{F}_{sup}$ indicates the projection network for supervised contrastive learning, $\mathcal{P}(i)=\{p\in\mathcal{A}|y_p=y_i\}$ is the index group for positive samples whose Koos grades are the same as the source sample $I^{(i)}$, $|\mathcal{P}(i)|$ refers to the number of samples in $\mathcal{P}(i)$.

\subsubsection{Koos grade prediction.}
By freezing the pre-trained $\mathbf{E}$ and $\mathbf{E}_\mathcal{H}$, we only fine-tune the final full connection layer with annotated real ceT1 and the corresponding generated hrT2 images. In this phase, the MSF-Koos-Net is trained with a cross-entropy loss.

\begin{figure}
    \centering
    \includegraphics[width=1\linewidth]{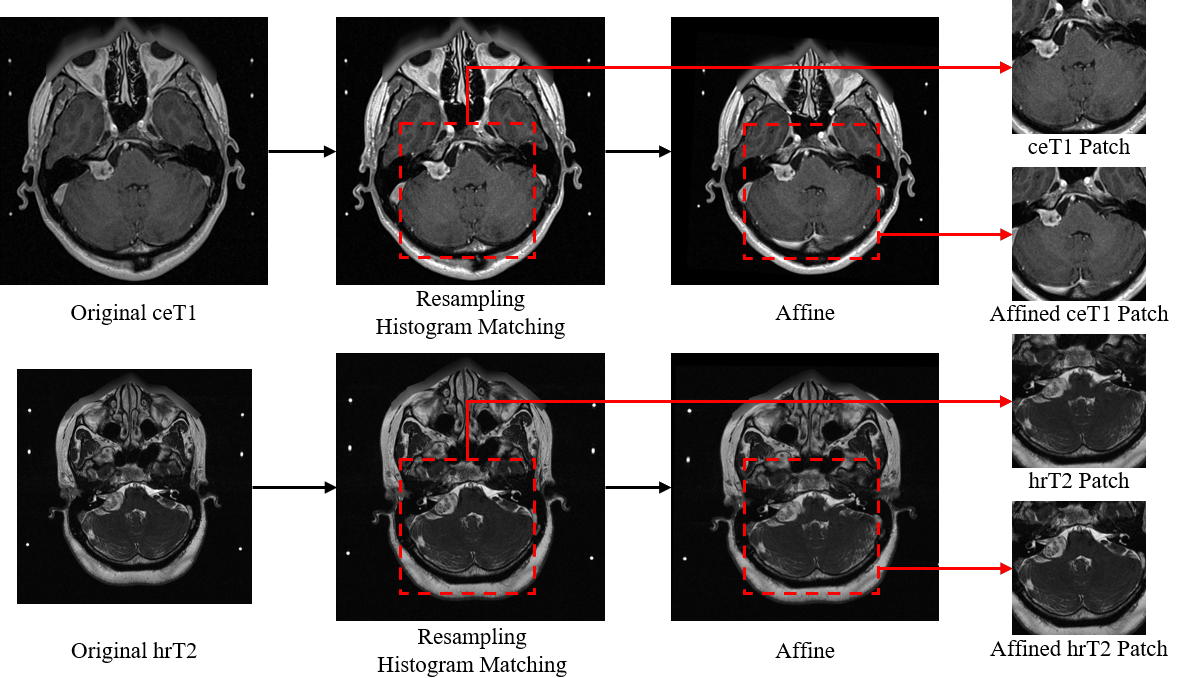}
    \caption{The pipeline of image preprocessing. Both ceT1 and hrT2 images are resampled and applied with histogram matching respectively. And then the images are registered to the same atlas by affine transformation. Finally, patches with the fixed region of interest are extracted from affined and non-affined images.}
    \label{fig:preprocess}
\end{figure}

\section{Experimental Results} \label{section:experiment}
\subsection{Materials and Implementation Details}
\subsubsection{Dataset.}
Training (210 subjects with ceT1 images and other unpaired 210 subjects with hrT2 images) and validation (64 subjects with hrT2 images) are datasets for the crossMoDA challenge, which is an extension of the publicly available Vestibular-Schwannoma-SEG collection released on The Cancer Imaging Archive (TCIA)~\cite{shapey2021segmentation,clark2013cancer,dorent2022crossmoda}.
All imaging datasets were manually segmented for cochlea and VS, and automated GIF parcellation masks are provided for the training source dataset.

\subsubsection{Data preprocessing.}
Figure~\ref{fig:preprocess} illustrates the pipeline of image preprocessing. All the images are first resampled to the spacing of $1\times0.4102\times0.4102$. Then we utilize histogram matching to normalize ceT1 and hrT2 images, separately. To improve the performance of domain adaptation, we select an identical hrT2 image as the atlas and employ intra- and inter-modality affine transformation on all the ceT1 and hrT2 images, respectively. Here, we utilize mutual information (MI) loss for ceT1 images and normalized cross-correlation (NCC) loss for hrT2 images. Finally, based on the distribution of tumor areas in the training set, we crop the images as the size of $80\times256\times256$ by setting a fixed region.
Limited by the device, we train the model in 2.5D mode -- adjacent three slices are treated as a three-channel 2D input.

\subsubsection{Implementation Details.}
We implemented our method using Pytorch with NVIDIA 3090 RTX. We optimized MSF-Net and MSF-Koos-Net with ADAM. MSF-Net is trained with a learning rate of $2\times10^{-4}$, a default of 1,000 epochs, and a batch size of 1. nnU-Net is trained with its default settings. MSF-Koos-Net is first pre-trained based on semi-supervised contrastive learning with a learning rate of $1\times10^{-2}$, a default of 100 epochs, and a batch size of 4. And then we fine-tune MSF-Koos-Net with a learning rate of $1\times10^{-4}$ and a default of 20 epochs.

\begin{figure*}[!htbp]
    \begin{minipage}{0.1\linewidth}
        \centerline{Real ceT1}
    \end{minipage}
    \begin{minipage}{0.14\linewidth}
        \centerline{\includegraphics[width=\textwidth]{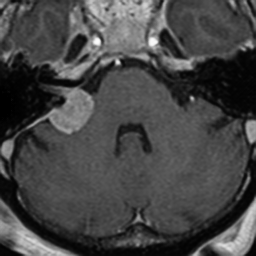}}
    \end{minipage}
    \begin{minipage}{0.14\linewidth}
        \centerline{\includegraphics[width=\textwidth]{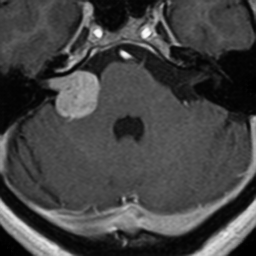}}
    \end{minipage}
    \begin{minipage}{0.14\linewidth}
        \centerline{\includegraphics[width=\textwidth]{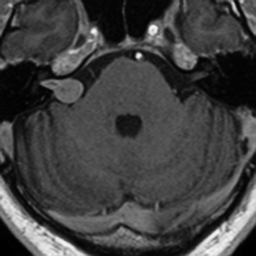}}
    \end{minipage}
    \begin{minipage}{0.14\linewidth}
        \centerline{\includegraphics[width=\textwidth]{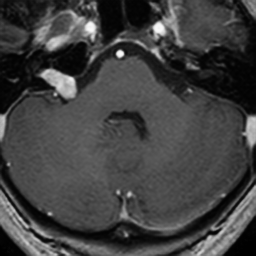}}
    \end{minipage}
    \begin{minipage}{0.14\linewidth}
        \centerline{\includegraphics[width=\textwidth]{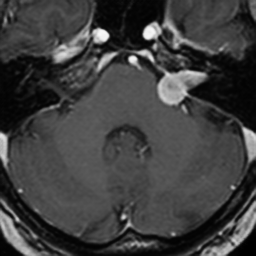}}
    \end{minipage}
    \begin{minipage}{0.14\linewidth}
        \centerline{\includegraphics[width=\textwidth]{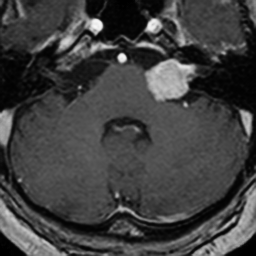}}
    \end{minipage}
    
    \vspace{2pt}
    
    \begin{minipage}{0.1\linewidth}
        \centerline{Gen hrT2}
    \end{minipage}
    \begin{minipage}{0.14\linewidth}
        \centerline{\includegraphics[width=\textwidth]{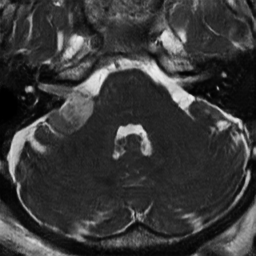}}
    \end{minipage}
    \begin{minipage}{0.14\linewidth}
        \centerline{\includegraphics[width=\textwidth]{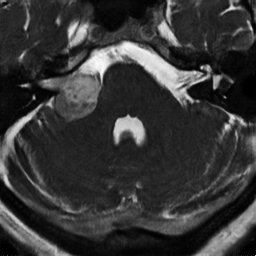}}
    \end{minipage}
    \begin{minipage}{0.14\linewidth}
        \centerline{\includegraphics[width=\textwidth]{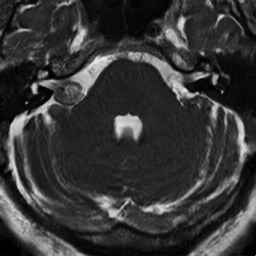}}
    \end{minipage}
    \begin{minipage}{0.14\linewidth}
        \centerline{\includegraphics[width=\textwidth]{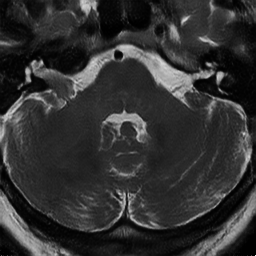}}
    \end{minipage}
    \begin{minipage}{0.14\linewidth}
        \centerline{\includegraphics[width=\textwidth]{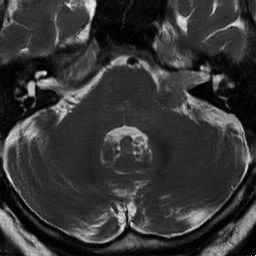}}
    \end{minipage}
    \begin{minipage}{0.14\linewidth}
        \centerline{\includegraphics[width=\textwidth]{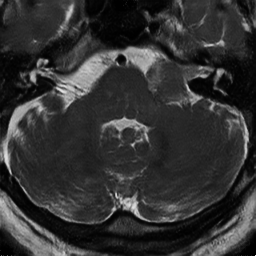}}
    \end{minipage}

    \begin{minipage}{0.1\linewidth}
        \centerline{}
    \end{minipage}
    \begin{minipage}{0.14\linewidth}
        \centerline{Case 1}
    \end{minipage}
    \begin{minipage}{0.14\linewidth}
        \centerline{Case 2}
    \end{minipage}
    \begin{minipage}{0.14\linewidth}
        \centerline{Case 3}
    \end{minipage}
    \begin{minipage}{0.14\linewidth}
        \centerline{Case 4}
    \end{minipage}
    \begin{minipage}{0.14\linewidth}
        \centerline{Case 5}
    \end{minipage}
    \begin{minipage}{0.14\linewidth}
        \centerline{Case 6}
    \end{minipage}
	
    \caption{Examples of real ceT1 images and corresponding generated hrT2 images from the proposed MSF-Net.} \label{fig:domainadaptation}
\end{figure*}

\subsection{Results}
Cross-modality domain adaptation results between ceT1 and hrT2 are shown in Fig.~\ref{fig:domainadaptation}. Real ceT1 images are correctly transferred to the hrT2 domain keeping the tumor structure unchanged.
Table~\ref{tab:seg} shows the segmentation results for the nnU-Net models training with fake hrT2 images generated by different methods. The proposed MSF-Net achieve better segmentation results than CycleGAN on the validation set, and the ablation study shows that adding the proxy task of VS and GIF can improve the performance of cross-modality domain adaptation.
For Koos grade prediction, the proposed MSF-Koos-Net achieves the best Macro-Average Mean Square Error (MAMSE) of 0.3940. MAMSE increases to 0.6805 when the pre-trained weights are not frozen. And further MAMSE increases to 0.8371 without pretraining with semi-supervised contrastive learning.

\begin{table}
\centering
\caption{Segmentation results for nnU-Net utilizing generated hrT2 images with different domain adaptation methods.}\label{tab:seg}
\begin{tabular}{|l|l|l|l|l|}
\hline
Methods & VS Dice $\uparrow$ & VS ASSD $\downarrow$ & Cochlea Dice $\uparrow$ & Cochlea ASSD $\downarrow$ \\
\hline
CycleGAN & 0.7402$\pm$0.2504 & 1.7556$\pm$5.3276 & 0.8202$\pm$0.0253 & 0.2325$\pm$0.1545 \\
MSF-Net w/o VS\&GIF & 0.7764$\pm$0.2025 & 0.6905$\pm$0.6437 & 0.8220$\pm$0.0510 & 0.3097$\pm$0.2986 \\
MSF-Net w/o GIF & 0.8288$\pm$0.0838 & 0.7901$\pm$1.0765 & 0.8285$\pm$0.0354 & 0.2507$\pm$0.1828 \\
MSF-Net & 0.8493$\pm$0.0683 & 0.5202$\pm$0.2288 & 0.8294$\pm$0.0268 & 0.2454$\pm$0.2102 \\
\hline
\end{tabular}
\end{table}

\section{Discussion}
In this study, we develop a cross-modality domain adaptation approach for VS and cochlea segmentation and also Koos grade prediction. In practice, our proposed MSF-Net is verified to convert the ceT1 domain to the hrT2 domain in an unsupervised manner. With a weight-shared encoder, MSF-Net is capable to learn joint multi-modality representation, given the ability of modality identification. The constraints on self-supervised modality recovery provide more structure consistency for the model training. Based on this, the proposed MSF-Net achieves better performance on domain adaptation than CycleGAN. This further affects the follow-up segmentation task, making the segmentation results of MSF-Net higher than that of CycleGAN. In addition, proxy tasks also have an important contribution to cross-modality domain adaptation. Segmentation of VS and brain structure can help less structural bias during image-to-image transformation and improve the segmentation accuracy.
Our proposed MSF-Koos-Net also achieves high accuracy in the cross-modality classification task.
Self-supervised medical image pretraining by contrastive learning has been proven to lead to Koos grade prediction performance improvements.
It is shown that freezing the pre-trained weights during finetuning stage of the model is effective for limited training data. That is to say, the number of parameters in deep learning-based classification models, especially 3D models, are too many for limited medical images. Weight-frozen strategy and contrastive learning are helpful in avoiding overfitting and capturing more representative information from images.
\subsubsection{Acknowledgement}
Luyi Han was funded by Chinese Scholarship Council (CSC) scholarship. This work was supported by the National Natural Science Foundation of China under Grant No. 62101365 and the startup foundation of Nanjing University of Information Science and Technology.

%
\bibliographystyle{splncs04}
\bibliography{refs}

\begin{thebibliography}{10}
\providecommand{\url}[1]{\texttt{#1}}
\providecommand{\urlprefix}{URL }
\providecommand{\doi}[1]{https://doi.org/#1}

\bibitem{bachman2019learning}
Bachman, P., Hjelm, R.D., Buchwalter, W.: Learning representations by
  maximizing mutual information across views. Advances in neural information
  processing systems  \textbf{32} (2019)

\bibitem{chen2020reusing}
Chen, R., Huang, W., Huang, B., Sun, F., Fang, B.: Reusing discriminators for
  encoding: Towards unsupervised image-to-image translation. In: Proceedings of
  the IEEE/CVF conference on computer vision and pattern recognition. pp.
  8168--8177 (2020)

\bibitem{choi2022using}
Choi, J.W.: Using out-of-the-box frameworks for contrastive unpaired image
  translation for vestibular schwannoma and cochlea segmentation: An approach
  for the crossmoda challenge. In: International MICCAI Brainlesion Workshop.
  pp. 509--517. Springer (2022)

\bibitem{clark2013cancer}
Clark, K., Vendt, B., Smith, K., Freymann, J., Kirby, J., Koppel, P., Moore,
  S., Phillips, S., Maffitt, D., Pringle, M., et~al.: The cancer imaging
  archive (tcia): maintaining and operating a public information repository.
  Journal of digital imaging  \textbf{26}(6),  1045--1057 (2013)

\bibitem{dong2021unsupervised}
Dong, H., Yu, F., Zhao, J., Dong, B., Zhang, L.: Unsupervised domain adaptation
  in semantic segmentation based on pixel alignment and self-training. arXiv
  preprint arXiv:2109.14219  (2021)

\bibitem{dorent2022crossmoda}
Dorent, R., Kujawa, A., Ivory, M., Bakas, S., Rieke, N., Joutard, S., Glocker,
  B., Cardoso, J., Modat, M., Batmanghelich, K., et~al.: Crossmoda 2021
  challenge: Benchmark of cross-modality domain adaptation techniques for
  vestibular schwnannoma and cochlea segmentation. arXiv preprint
  arXiv:2201.02831  (2022)

\bibitem{he2020momentum}
He, K., Fan, H., Wu, Y., Xie, S., Girshick, R.: Momentum contrast for
  unsupervised visual representation learning. In: Proceedings of the IEEE/CVF
  conference on computer vision and pattern recognition. pp. 9729--9738 (2020)

\bibitem{huang2021gloria}
Huang, S.C., Shen, L., Lungren, M.P., Yeung, S.: Gloria: A multimodal
  global-local representation learning framework for label-efficient medical
  image recognition. In: Proceedings of the IEEE/CVF International Conference
  on Computer Vision. pp. 3942--3951 (2021)

\bibitem{isensee2021nnu}
Isensee, F., Jaeger, P.F., Kohl, S.A., Petersen, J., Maier-Hein, K.H.: nnu-net:
  a self-configuring method for deep learning-based biomedical image
  segmentation. Nature methods  \textbf{18}(2),  203--211 (2021)

\bibitem{khosla2020supervised}
Khosla, P., Teterwak, P., Wang, C., Sarna, A., Tian, Y., Isola, P., Maschinot,
  A., Liu, C., Krishnan, D.: Supervised contrastive learning. Advances in
  Neural Information Processing Systems  \textbf{33},  18661--18673 (2020)

\bibitem{meyer2020improving}
Meyer, J., Eitel, A., Brox, T., Burgard, W.: Improving unimodal object
  recognition with multimodal contrastive learning. In: 2020 IEEE/RSJ
  International Conference on Intelligent Robots and Systems (IROS). pp.
  5656--5663. IEEE (2020)

\bibitem{shapey2021segmentation}
Shapey, J., Kujawa, A., Dorent, R., Wang, G., Dimitriadis, A., Grishchuk, D.,
  Paddick, I., Kitchen, N., Bradford, R., Saeed, S.R., et~al.: Segmentation of
  vestibular schwannoma from mri, an open annotated dataset and baseline
  algorithm. Scientific Data  \textbf{8}(1), ~1--6 (2021)

\bibitem{shin2022cosmos}
Shin, H., Kim, H., Kim, S., Jun, Y., Eo, T., Hwang, D.: Cosmos: Cross-modality
  unsupervised domain adaptation for 3d medical image segmentation based on
  target-aware domain translation and iterative self-training. arXiv preprint
  arXiv:2203.16557  (2022)

\bibitem{wu2018unsupervised}
Wu, Z., Xiong, Y., Yu, S.X., Lin, D.: Unsupervised feature learning via
  non-parametric instance discrimination. In: Proceedings of the IEEE
  conference on computer vision and pattern recognition. pp. 3733--3742 (2018)

\bibitem{yuan2021multimodal}
Yuan, X., Lin, Z., Kuen, J., Zhang, J., Wang, Y., Maire, M., Kale, A., Faieta,
  B.: Multimodal contrastive training for visual representation learning. In:
  Proceedings of the IEEE/CVF Conference on Computer Vision and Pattern
  Recognition. pp. 6995--7004 (2021)

\bibitem{zhu2017unpaired}
Zhu, J.Y., Park, T., Isola, P., Efros, A.A.: Unpaired image-to-image
  translation using cycle-consistent adversarial networks. In: Proceedings of
  the IEEE international conference on computer vision. pp. 2223--2232 (2017)

\end{thebibliography}
	
\end{document}